\documentclass[journal,twoside,web]{ieeecolor}
\usepackage{generic}
\usepackage{cite}
\usepackage{amsmath,amssymb,amsfonts}
\usepackage{algorithmic}
\usepackage{graphicx}
\usepackage{epstopdf}
\usepackage{algorithm,algorithmic}
\usepackage{hyperref}
\hypersetup{hidelinks}
\usepackage{textcomp}
\def\BibTeX{{\rm B\kern-.05em{\sc i\kern-.025em b}\kern-.08em
    T\kern-.1667em\lower.7ex\hbox{E}\kern-.125emX}}
\begin{document}
\title{Clinically-Inspired Hierarchical Multi-Label Classification of Chest X-rays with a Penalty-Based Loss Function}
\author{Mehrdad Asadi, Komi Sodoké, Ian J. Gerard, and Marta Kersten-Oertel
\thanks{This work has been submitted to the IEEE for possible publication. Copyright may be transferred without notice, after which this version may no longer be accessible.}
\thanks{This work was funded by an NSERC Discovery Grant (RGPIN-2017-06722).}
\thanks{Mehrdad Asadi is with Gina Cody School of Engineering and Computer Science, Concordia University, Montreal, QC, Canada (email: mehrdad.asadi@mail.concordia.ca)}
\thanks{Marta Kersten-Oertel is with Gina Cody School of Engineering and Computer Science, Concordia University, Montreal, QC, Canada (e-mail: marta.kersten@concordia.ca)}
\thanks{Komi Sodoké is with YULCOM Technologies, Montreal, QC, Canada (e-mail: skomi@yulcom.ca)}
\thanks{Ian J. Gerard is with the Division of Radiation Oncology, McGill University Health Centre, Montreal, QC, Canada (e-mail: ian.gerard@mail.mcgill.ca)}}
\maketitle

\begin{abstract}
In this work, we present a novel approach to multi-label chest X-ray (CXR) image classification that enhances clinical interpretability while maintaining a streamlined, single-model, single-run training pipeline. Leveraging the CheXpert dataset and VisualCheXbert-derived labels, we incorporate hierarchical label groupings to capture clinically meaningful relationships between diagnoses. To achieve this, we designed a custom hierarchical binary cross-entropy (HBCE) loss function that enforces label dependencies using either fixed or data-driven penalty types. Our model achieved a mean area under the receiver operating characteristic curve (AUROC) of 0.903 on the test set. Additionally, we provide visual explanations and uncertainty estimations to further enhance model interpretability. All code, model configurations, and experiment details are made available.
\end{abstract}


\section{Introduction}
\label{sec:introduction}
\IEEEPARstart{I}{n} the field of medical imaging, multi-label classification is important for diagnosing a wide range of conditions from an image set, especially in 2D radiographic modalities like chest X-rays (CXR). CXR 
analysis poses unique challenges due to the complex and overlapping nature of thoracic diseases, where capturing clinically meaningful label dependencies is important to improve diagnostic reliability and efficiently rule in or out diagnoses that may require urgent intervention. 

Lung pathologies remain one of the leading causes of morbidity and mortality worldwide~\cite{Dattani2023Causes}, highlighting the significant role of accurate and timely radiographic interpretation. However, diagnostic variability persists due to differences in radiologists' experience and expertise, leading to inconsistent interpretations, particularly in subtle or complex cases. In addition, long working hours can impair concentration, increasing the likelihood of diagnostic errors~\cite{Kasalak2023Workload}. The increasing demand for imaging studies, coupled with a global shortage of trained radiologists, particularly in low-resource settings, has significantly increased radiology workloads~\cite{Bruls2020Workload}. This strain often results in delays in diagnosis, where heavy workloads and resource constraints compromise timely interpretation, negatively impacting clinical outcomes. Despite the advancement of Deep Learning (DL) techniques, many current approaches struggle to incorporate clinical hierarchies effectively. This limitation is particularly evident in hierarchical classification models that do not sufficiently leverage clinical insights through penalties or relationships between parent and child labels, thus limiting both interpretability and performance.

To address these challenges, we introduce a hierarchical multi-label classification framework for CXRs, which organizes clinically related labels into parent-child relationships. In this framework, we propose a novel hierarchical binary cross-entropy (HBCE) loss function that applies penalties when child labels are predicted as positive without corresponding positive predictions for their parent labels. To optimize the effectiveness of this loss function, we explore two key penalty strategies: a fixed penalty approach and a data-driven penalty method. The data-driven method adjusts the penalty based on the likelihood of parent-child label dependencies, with a range of scale factors to fine-tune the impact of the penalty according to the strength of these relationships.

The primary objective of this work is to improve the clinical interpretability of CXR classifications by using hierarchical groupings that mirror clinical decision-making processes: often, the most dangerous and/or life-threatening diagnoses that require immediate and urgent intervention must be ruled out before further workup for less severe diagnoses is performed. While a CXR provides less information than three-dimensional tomographic imaging, they are a fundamental tool used in many inpatient hospital settings due to the portability of imaging units. Many severely ill and unstable patients often undergo a CXR in their hospital bed before more complex imaging is performed to minimize the risk of unnecessary manipulation that could decompensate their fragile health status. These CXRs are often reviewed by non-radiology trained medical professionals and can be challenging to interpret. By introducing a custom hierarchical loss function we aim to improve the model performance and explainability for transparency on diagnosis determination. By systematically evaluating the hierarchy and penalty types, we comprehensively analyze their effects on predictive performance. We demonstrate the practical benefits of leveraging clinical insights for accurate and interpretable multi-label classification.

The proposed framework achieved a weighted AUROC of 0.9034 on the CheXpert dataset using a single-model, single-run pipeline, showing the efficacy of the hierarchical structure and the custom HBCE loss function. Data-driven penalties showed the potential to improve predictive accuracy, while visual explanations and uncertainty estimations enhanced model interpretability and transparency. 

To promote transparency and reproducibility, all code, model configurations, and experiment details have been made available in a public Git repository: \underline{\href{https://github.com/the-mercury/CIHMLC.git}{CIHMLC}}.

\section{Related Work}

\subsection{Multi-Label Classification in Medical Imaging}
Multi-label classification plays an important role in analyzing CXRs, particularly when handling a diverse range of pathologies that overlap or present concurrently. Prior studies have focused on improving the performance of such models. For example, Wang \textit{et al.}~\cite{wang2024awareness} introduced CXR×MLAGCPL, a multi-label classification model for CXRs that leverages both local label correlations and global co-occurrence patterns for improved disease prediction. The model captures nuanced inter-pathological patterns by combining a Local Awareness Module (LAM) for image-specific label dependencies and a Global Co-occurrence Priori Learning (GCPL) module for dataset-wide label relationships. Evaluated on large datasets, CXR×MLAGCPL achieved a good performance on most labels (mean AUROC of 0.805 and 0.810 on all 14 and common 5 classes of the CheXpert~\cite{irvin2019chexpert} dataset respectively), highlighting the benefit of jointly considering local and global dependencies in multi-label medical image classification.

A similar study by Zhang \textit{et al.}~\cite{zhang2024discriminative}, proposed the Label Correlation Guided Discriminative Label Feature Learning (LCFL) model for CXR classification which uses a self-attention-based Label Correlation Learning (LCL) module to capture global label correlations and a Discriminative Label Feature Learning (DLFL) module for feature enhancement through label-level contrastive learning. This framework allows the model to learn distinctive label-specific features, yielding a mean AUROC score of 0.764 on the CheXpert dataset and U-zeros setting, utilizing both global and local label correlations to guide discriminative feature learning.

CvTGNet by Lu \textit{et al.}~\cite{lu2024CvTGNet} integrated convolutional and transformer architectures alongside graph-based co-occurrence modeling to enhance the multi-label classification of CXRs. By combining the strengths of Convolutional Vision Transformers for spatial detail extraction and Graph Convolutional Networks for pathological relationship learning, the model effectively captures both image-specific and inter-label dependencies, achieving an overall AUROC score of 0.840 across all 14 CheXpert pathologies.

In another study, Liu \textit{et al.}~\cite{liu2023Global} introduced ML-LGL. This framework enhances multi-label CXR classification by applying a clinical-inspired curriculum learning strategy, gradually training the model from common to rare abnormalities. This approach integrates three selection functions—correlation, similarity, and frequency-based functions—to build a radiologist-like curriculum, achieving a mean AUROC of 0.889 and 0.841 for 5 and 13 labels respectively on the CheXpert dataset and U-zeros setting.

On the other hand, the Semantic Similarity Graph Embedding (SSGE) framework by Chen \textit{et al.}~\cite{chen2022Graph} enhances multi-label CXR classification by embedding semantic relationships across images, using a “Teacher-Student” model. By combining a similarity graph with CNN-based feature extraction and GCN-based feature recalibration, the model achieves consistent, semantically informed features, leading to an overall AUROC score of 0.836 on the CheXpert dataset and U-zeros setting.

CheXtransfer by Ke \textit{et al.}~\cite{ke2021CheXtransfer} investigates the transferability of ImageNet-pretrained CNNs for CXR classification by evaluating 16 popular CNN architectures, analyzing relationships between model size, parameter efficiency, and classification performance, finding no correlation between ImageNet and CheXpert performance but significant gains from ImageNet pretraining. The study demonstrates that truncating layers can improve parameter efficiency without compromising accuracy, enhancing both performance and interpretability in medical imaging models.

The similar study by Huang \textit{et al.}~\cite{huang2022Transfer} leverages transfer learning for multilabel CXR classification, utilizing CNNs and three source datasets to analyze fine-tuning, layer transfer, and combined transfer approaches. Their results show that although initializing the model with ImageNet weights had the best training performance, it performed worse in the test process. They also showed integrating related datasets (e.g., ChestX-ray14~\cite{wang2017ChestX} and CheXpert) improves training accuracy but it does not achieve the best performance on the test set.

Recent advancements in multi-label CXR classification emphasize integrating spatial feature extraction with label dependency modeling to enhance diagnostic accuracy. Methods leveraging hybrid architectures, semantic embedding, and curriculum learning effectively capture both image-specific and label co-occurrence patterns, reflecting a shift toward more accurate and interpretable classification models.

\subsection{Hierarchical Learning in Medical Imaging}
The use of hierarchical networks has introduced a significant advancement, particularly with the introduction of Hierarchical Multi-Label Classification Networks (HMCN). This architecture represents one of the first attempts to explicitly incorporate hierarchical structures within deep neural network frameworks for multi-label classification~\cite{Wehrmann2018Hierarchical}. Compared to traditional flat multi-label approaches, HMCNs integrate hierarchical relationships directly into the learning process, leading to accurate predictions that align with the structure of the label space. This is particularly relevant in domains where hierarchical dependencies are naturally present, such as medical imaging, where disease categories often share anatomical or pathological relationships.

In work by Chen \textit{et al.}~\cite{Chen2020hierarchical} a two-stage Hierarchical Multi-Label Classification (HMLC) approach is introduced for CXR analysis. It utilizes clinically relevant taxonomies to enhance interpretability and manage incomplete labeling common in medical datasets. The method first trains the model to predict conditional probabilities within the label hierarchy, focusing on sibling categories, before fine-tuning with a numerically stable cross-entropy loss function to derive unconditional probabilities, thereby improving classification stability and performance. Evaluated on the PLCO~\cite{Aberle2000plco} and PadChest~\cite{padchest2020} datasets, the approach outperformed traditional flat classifiers and other hierarchical models, achieving the highest AUROC (0.887) reported on PLCO and showing resilience to missing labels. This study demonstrates HMLC’s utility in producing clinically aligned, interpretable predictions, offering a significant advancement for CXR computer-aided diagnosis (CAD) and suggesting broader applications across hierarchical classification tasks in medical imaging.

Similarly, the study by Pham \textit{et al.}~\cite{Pham2021hierarchical} presents a deep learning-based framework for multi-label CXR classification that integrates hierarchical disease dependencies and addresses label uncertainty, advancing upon previous work in the field. By using a conditional training process based on a predefined disease hierarchy, the model learns relationships among parent and child disease labels, allowing it to make clinically consistent predictions. To manage uncertain labels, the authors apply Label Smoothing Regularization (LSR), reducing model overconfidence in cases of label ambiguity. Trained on the CheXpert dataset, their combined Conditional Training (CT) and LSR in the U-zeros setting achieved an AUROC score of 0.884 on the five CheXpert labels. In addition, they evaluated an ensemble of six CNN architectures and achieved an AUROC of 0.940, outperforming prior methods and even surpassing most radiologists on the validation set. Their method highlights the value of incorporating hierarchical relationships and uncertainty handling in improving the interpretability and accuracy of automated CXR analysis.

Hierarchical Multi-Label Classification Networks (HMCNs) and related approaches represent significant advancements by incorporating hierarchical label structures directly into the learning process, leading to clinically interpretable and accurate predictions.  Table~\ref{table:abalation} presents the results of these studies evaluated on the CheXpert dataset compared to ours. 

\section{Methodology}
\subsection{Novelty and Divergence from Prior Work}
The proposed method emphasizes enhancing the clinical interpretability and explainability of model predictions while maintaining strong performance on the CheXpert dataset. This highlights a new approach for integrating domain knowledge with DL in a clinically meaningful way. We developed a penalty-based loss function that enforces consistency between child and parent labels, addressing clinically implausible predictions, a feature not commonly seen in traditional dependency models (e.g., those using label co-occurrence or graph convolutional networks). Moreover, while hierarchical models use multi-stage training to refine parent and child label relationships, this study opts for a single-run pipeline, simplifying the process and reducing computational complexity but potentially limiting nuanced adjustments for hierarchical learning.

\subsection{Dataset and Labeling Strategy}
\begin{figure*}[ht]
    \centering
    \includegraphics[width=0.9\textwidth]{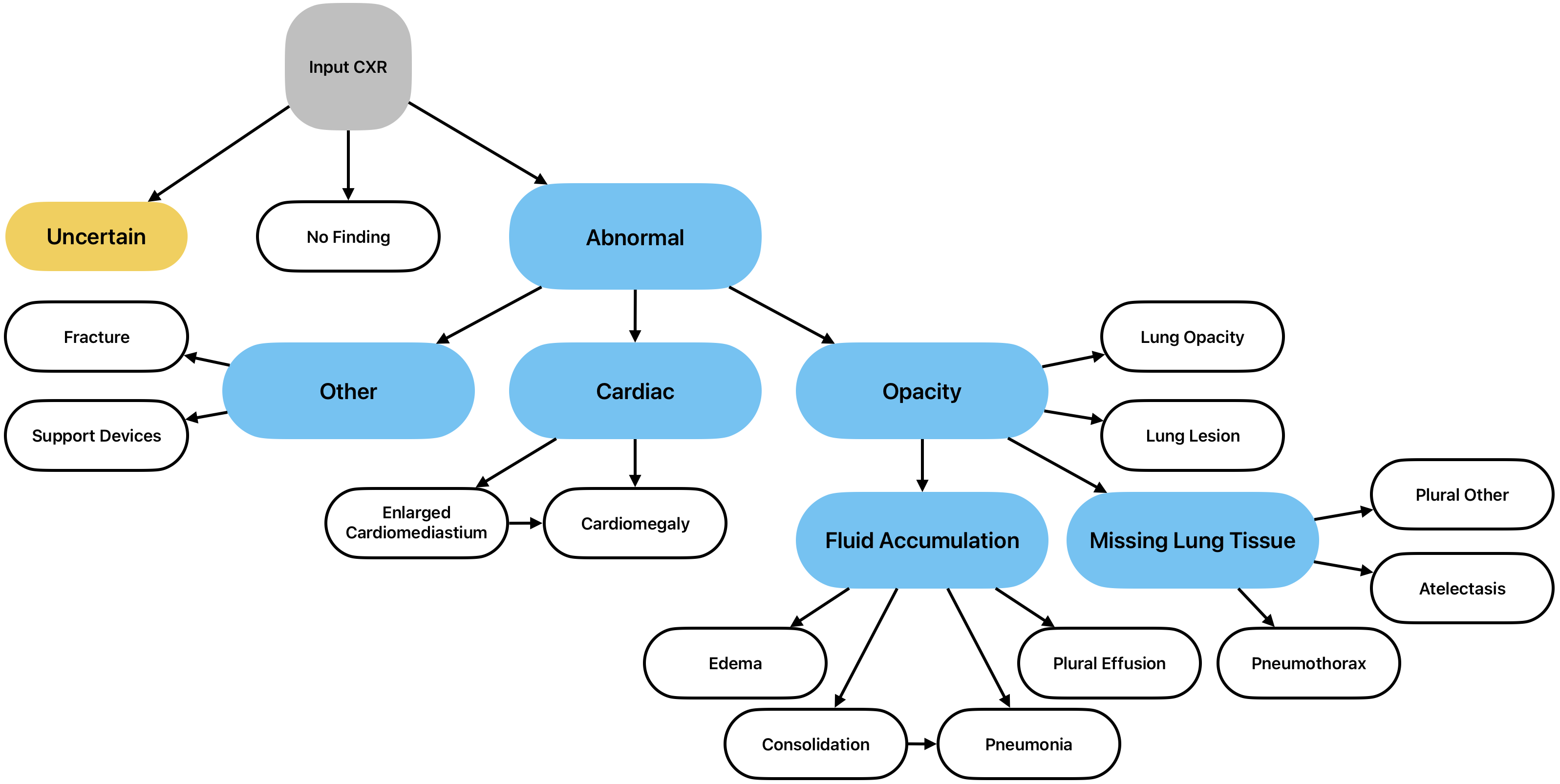}
    \caption{Clinically-Inspired Taxonomy}
    \label{fig:H4}
\end{figure*}
We used the CheXpert dataset, one of the largest publicly available CXR datasets, which consists of over 224,000 CXRs from more than 65,000 patients. CheXpert has become a standard benchmark for multi-label classification in medical imaging due to its diverse range of labeled thoracic pathologies, including atelectasis, cardiomegaly, pulmonary consolidation, pulmonary edema, and pneumonia, among others. This dataset provides annotations derived from radiologist reports using rule-based labelers, specifically VisualCheXbert~\cite{jain2021visualchexbert}, compared to the CheXpert and CheXbert~\cite{smit2020chexbert} labelers.

However, directly using these labels poses limitations due to the complex interdependence between the thoracic pathologies. To address this, we introduced hierarchical label groupings where individual pathologies are grouped under newly defined parent categories based on clinical insights. For instance, labels such as Pleural Effusion and Pneumonia, Edema, and Consolidation were grouped under the parent label “Fluid Accumulation,” capturing clinical dependencies in the underlying pathologies~\cite{webmd_pleural_effusion, mayo_clinic_pneumonia, cleveland_clinic_pulmonary_edema, radiopaedia_}. Furthermore, we labeled all instances with neither positive findings nor "No Finding" as “Uncertain,” which allowed us to capture a broader spectrum of clinical uncertainty in the dataset.

This hierarchical grouping, informed by previous works~\cite{Chen2020hierarchical, smithuis2014lungdisease} and feedback from a clinician, ensures that the model leverages clinical knowledge and reflects real-world diagnostic relationships, thereby potentially improving the predictive performance and, more importantly, clinical interpretability of the model.

\subsubsection{Clinical Relevance of the Groupings}
The hierarchy’s structure in Fig~\ref{fig:H4} reflects the relationships between different conditions, making it possible for the model to capture clinical dependencies and associations more effectively. The structure aligns with clinical reasoning as follows:
\begin{itemize}
    \item \textbf{Grouping Related Conditions:} By categorizing labels into clinically meaningful groups, the hierarchy better relates different diagnoses based on visual features of the CXR.  For example, the appearance of an enlarged cardiac silhouette or altered contours in the mediastinum is often a suggestion of a primary cardiac diagnosis. In contrast, gravity-dependent pulmonary opacifications often suggest fluid infiltration and may include a spectrum of pathologies. These are frequently described as hazy or opaque regions and could indicate pleural effusions, pulmonary edema, or fluid within the parenchymal spaces depending on their geometric distribution within the CXR.
	\item \textbf{Parent-Child Dependencies:} In clinical practice, some conditions naturally lead to others, i.e., consolidations could precede pneumonia~\cite{radiopaedia_}. By structuring the hierarchy this way, the model can potentially better understand and reflect these dependencies in its predictions.
	\item \textbf{Improved Interpretability and Clinical Decision Making:} The hierarchical approach makes the model’s predictions easier to interpret by clinicians~\cite{sen2021hierarchical} as it categorizes pathologies into broader groups (e.g., lung opacities vs. cardiac abnormalities), then refining the diagnosis based on specific child labels (e.g., Cardiomegaly under Cardiac). Similar to clinical workflows, broad diagnoses are initially considered before narrowing down to more specific conditions that may require more resource-intensive or invasive procedures to accurately diagnose. Additionally, it may help radiologists and/or other interpreting clinicians validate whether the clinical status of the patient being evaluated is congruent with the findings seen on imaging.
	\item \textbf{Capturing Clinical Uncertainty:} The use of an Uncertain category allows the model to acknowledge cases where there isn’t a clear diagnosis. This is an important insight for clinical decision-making as it may indicate that higher-quality imaging, such as computed tomography, may be necessary. Additionally, it will help reduce the rates of false positives where applying the best-matched diagnosis that may be incorrect when a certainty threshold is not reached when analyzing ambiguous CXRs. This feature thus makes the model more practical for use in diagnostic support.
\end{itemize}

\subsection{Model Architecture}
We used DenseNet121~\cite{Huang2017DenseNet} as the base architecture due to its widespread use in medical image analysis and its dense connectivity pattern, which mitigates the vanishing gradient problem, promotes feature reuse, and enables efficient parameter utilization. To adapt DenseNet121 for a multi-label classification task, we extended the architecture with several additional layers:
\begin{itemize}
    \item A \textit{Conv2D} layer with 512 filters was added after the DenseNet121 backbone to enhance feature extraction from CXR images, specifically targeting finer details related to the hierarchical structure of pathologies.
	\item \textit{Batch Normalization} was applied to stabilize and accelerate the training process.
	\item A \textit{Global Average Pooling (GAP)} layer was used instead of a fully connected layer to reduce overfitting while maintaining the critical spatial features in the images.
    \item A fully connected dense layer with 128 neurons and a \textit{ReLU} activation function is applied to introduce non-linearity and enable the model to capture complex interactions between the high-level features extracted from the \textit{GAP} layer.
	\item \textit{Dropout} layers were used to prevent over-fitting by randomly dropping connections during training with a dropout rate of {0.5}.
	\item Finally, a \textit{Dense} layer with a \textit{sigmoid} activation function was used to output the probability scores for each label, enabling multi-label predictions.
\end{itemize}

The model was initialized with random weights and trained from scratch, as the domain-specific nature of CXR images often benefits from task-specific training~\cite{raghu2019transfusion}. 

\subsection{Clinically-Inspired Hierarchical Loss Function}
We developed a loss function that aims to reflect clinically meaningful dependencies between labels. In multi-label classification tasks such as CXR analysis, individual labels are not independent of one another. For instance, the presence of a specific pathology (e.g., Pleural Effusion) can increase the likelihood of another pathology (e.g., Pneumonia). To capture these dependencies, our hierarchical loss function incorporates penalties when the model predicts child labels inconsistently with their corresponding parent labels.

\subsubsection{Binary Cross-Entropy}
At the core of our loss function is the Binary Cross-Entropy (BCE) Loss, which is traditionally used in multi-label classification. BCE measures the difference between the predicted probabilities $y_{\text{pred}}$ and the true binary labels $y_{\text{true}}$ for each label. For  $L$  labels and a batch size of $B$, the BCE loss is given by:

\begin{equation}
\begin{split}
L_{\text{BCE}}(y_{\text{true}}, y_{\text{pred}}) = \\ 
- \frac{1}{B}\sum_{b=1}^{B} \sum_{l=1}^{L} \Big( &y_{\text{true}}^{(b, l)} \log(y_{\text{pred}}^{(b, l)}) + (1 - y_{\text{true}}^{(b, l)}) \log(1 - y_{\text{pred}}^{(b, l)}) \Big)
\end{split}
\end{equation}

While effective for independent labels, this loss function fails to capture the hierarchical dependencies that are clinically relevant in CXR analysis.

\subsubsection{Hierarchical Penalty for Parent-Child Relationships}
Incorporating a hierarchical structure requires addressing parent-child dependencies in label predictions. For example, if a parent label (e.g., “Fluid Accumulation”) is predicted as negative, it is inconsistent for a child label (e.g., “Pleural Effusion”) to be predicted as positive. To enforce such consistency, a penalty term was added to the standard BCE loss. This penalty is designed to increase the loss by a factor to discourage clinically implausible predictions.

Let  $P_{p,c}$  represent the penalty applied between a parent label $p$  and its child label $c$. The hierarchical penalty for each pair of parent and child labels is computed as:

\begin{equation}
P_{p,c} = \text{Penalty}(p, c) \cdot \mathbf{1}\{ y_{\text{pred},p} < 0.5 \text{ and } y_{\text{pred},c} > 0.5 \}
\end{equation}

\noindent where:
\begin{itemize}
    \item $y_{\text{pred},p}$  is the predicted probability for the parent label,
	\item $y_{\text{pred},c}$  is the predicted probability for the child label,
	\item $\mathbf{1}\{\cdot\}$  is an indicator function that triggers the penalty when the condition holds. 
\end{itemize}

\subsubsection{Fixed vs. Data-Driven Penalties}
The hierarchical loss function operates in two modes depending on the source of the penalties:
\begin{itemize}
    \item Fixed Penalty: A constant penalty value is assigned to all parent-child inconsistencies. This penalty mode is straightforward and computationally efficient, but it lacks adaptability to data:
        \begin{equation}
            \text{Penalty}(p,c) = \beta
        \end{equation}
    
	\item Data-Driven Penalty: The penalties are dynamically computed based on the empirical likelihood of child labels given the parent labels in the training dataset. This approach introduces a degree of adaptiveness and should better reflect the relationships between labels.
\end{itemize}

In the data-driven approach, the penalty  $\text{Penalty}(p,c)$  is calculated as:

\begin{equation}
\text{Penalty}(p,c) = \frac{N_{\text{parent}_p = 0, \text{child}c = 1} + \epsilon}{N_{\text{parent}_p = 0} + 2\epsilon}
\end{equation}

\noindent where:
\begin{itemize}
    \item $N_{\text{parent}_p = 0, \text{child}_c = 1}$ is the count of instances where the parent label $p$ is negative and the child label $c$ is positive,
	\item $N_{\text{parent}_p = 0}$ is the total count of instances where the parent label $p$ is negative,
	\item $\epsilon$ is a small Laplace smoothing factor to avoid division by zero.
\end{itemize}

The total hierarchical penalty is scaled by a scale factor  $\lambda$, which controls the strength of the hierarchical penalty relative to the BCE loss. The final total HBCE loss is expressed as:

\begin{equation}
L_{\text{HBCE}} = L_{\text{BCE}} + \lambda \sum_{p,c} P_{p,c}
\end{equation}

\subsection{Training Strategy}
The model was trained using the Adam optimizer, with an initial learning rate of 0.0001. Following a learning rate reduction-on-plateau strategy, this learning rate decreased by a factor of 0.9 after the validation loss plateaued. Early stopping was used to prevent overfitting and halting training if the validation loss did not decrease for three consecutive epochs. To ensure robust model performance, we used checkpointing, saving the model when either the validation loss decreased or the AUROC increased. 

Additionally, we increased the input image size to 320x320 to capture higher-resolution features. Given the increased input size and memory limitations, the batch size was set to 16, balancing computational constraints with performance needs and increasing generalizability~\cite{smith2018disciplined}. The data splits are as follows: \textit{Train}: $223,414$, \textit{Validation}: $234$, and \textit{Test}: $668$. Each subset is mutually exclusive, without overlapping data points, ensuring that the model performance metrics are unbiased and accurately reflect its generalization capabilities.

We also incorporated random image augmentations using TensorFlow’s image processing library. The augmentations included horizontal and vertical flips, random brightness adjustments with a delta of 0.1, and random contrast variations with lower and upper bounds of 0.9 and 1.1, respectively. Notably, while the seed for random number generation was fixed to ensure reproducibility across batches, the augmentations were allowed to vary, thus enhancing the model’s generalization capabilities.

The training was performed on an \textit{Ubuntu 20.04.2 LTS (Focal Fossa)} with an \textit{NVIDIA GeForce RTX 2080 Ti} GPU and an \textit{Intel Core i9-9900K} CPU @ \textit{3.60GHz}. \textit{Tensorflow} and \textit{Keras} versions were \textit{2.16.1} and \textit{3.0.5}, respectively.

\subsection{Monte Carlo Uncertainty Estimation}
To enhance the transparency and interpretability of the model, we used Monte Carlo dropout during inference, a technique that enables uncertainty estimation by making multiple forward passes through the model with active dropout layers. This yields a distribution of predictions, from which we can compute the mean and standard deviation for each label, quantifying the confidence of the model in its predictions. These statistics provide a measure of the uncertainty associated with each prediction, providing clinicians with information on the reliability of the model results.




\subsection{Class Activation Maps}
To further facilitate interpretability and explainability, we derived Class Activation Maps (CAM) using the Grad-CAM method~\cite{selvaraju2017gradcam}. This technique generates heatmaps indicating the regions of the input image that most strongly influence the model’s predictions. By backpropagating the gradients of the target class for the final convolutional layer, we obtained visual explanations that highlight the relevant areas of the CXR contributing to each predicted label. Such visualizations are important for enhancing the model’s clinical applicability, as they allow practitioners to verify whether the model’s attention aligns with the expected regions of interest or abnormalities seen during physical exam i.e. auscultation of the lungs and/or heart. 

\section{Results}
\begin{table*}[ht]
    \centering
    \begin{tabular}{l c | c c c c c c c}
        \hline
        \textbf{Penalty Type} & \textbf{Scale Factor} &\textbf{Atelectasis} & \textbf{Cardiomegaly} & \textbf{Consolidation} & \textbf{Edema} & \textbf{Pleural Effusion} & \textbf{Mean} \\
        \hline
        \textit{N/A} | without ``Uncertain'' Label & \textit{N/A}  & 0.878 & 0.856 & 0.900 & 0.893 & \textbf{0.950} & 0.895 \\
        \textit{N/A} | with ``Uncertain'' Label & \textit{N/A} & \textbf{[0.883]} & \textbf{[0.861]} & \textbf{[0.903]} & \textbf{[0.905]} & 0.944 & \textbf{[0.899]} \\
        \hline
        Data-Driven & 0.3 & 0.877 & 0.839 & 0.868 & 0.886 & 0.928 & 0.880 \\
        Data-Driven & 0.5 & \textbf{0.879} & \textbf{0.853} & \textbf{0.880} & \textbf{0.901} & \textbf{[0.945]} & \textbf{0.892} \\
        Data-Driven & 0.7 & 0.874 & 0.849 & 0.868 & 0.895 & 0.917 & 0.881 \\
        Data-Driven & 1.0 & 0.873 & 0.852 & 0.875 & 0.894 & 0.933 & 0.885 \\
        \hline
        Fixed & 0.3 & 0.765 & \textbf{0.855} & 0.883 & 0.886 & \textbf{0.942} & \textbf{0.888} \\
        Fixed & 0.5 & 0.704 & 0.841 & 0.875 & 0.884 & 0.926 & 0.877 \\
        Fixed & 0.7 & \textbf{0.783} & 0.850 & \textbf{0.898} & 0.881 & 0.930 & 0.885 \\
        Fixed & 1.0 & 0.760 & 0.849 & 0.892 &\textbf{ 0.890} & 0.930 & 0.885 \\
        \hline
    \end{tabular}
    \caption{AUROC Comparison with Different Strategies on 5 CheXpert Pathologies Evaluated on the Official CheXpert Test Set. The highest AUROC is highlighted in \textbf{BOLD} for each column and penalty strategy. The Overall Maximum for each column is in brackets (\textbf{[]}). All penalty-based scenarios include the ``Uncertain'' label. $\beta = 1$ for all fixed penalty strategies.
    }
    \label{table:comparison_5_CheXpert_Pathologies}
\end{table*}

The experiments were initiated by training two baseline models with and without the ``Uncertain'' label to assess the effect of its inclusion. Table~\ref{table:comparison_5_CheXpert_Pathologies} provides the AUROC results across various training strategies on five common pathologies from the CheXpert dataset, along with the influence of the ``Uncertain'' label. It examines two primary setups: flat training with and without the ``Uncertain'' label, and hierarchical training with data-driven and fixed penalties at varying scale factors. Given that the correlation between parent and child labels ranges from 0.814 to 0.999, a fixed penalty value of 1 was selected to ensure a fair comparison of different penalty strategies. It is worth noting the flat training with the Uncertain label achieves the highest mean AUROC (0.899), with superior performance in several pathologies such as Atelectasis (0.883), Cardiomegaly (0.861), Consolidation (0.903), and Edema (0.905). In addition, flat training shows significantly better performance on Cardiomegaly, Consolidation, and Pleural Effusion, Pneumonia, and marginally better on the five pathologies mean ($p = 0.0484$, $0.027$, $0.0473$, $0.0308$, $0.0057$, respectively).

\subsection{Uncertain Label}
To evaluate the significance of including the ``Uncertain'' label on model performance, a paired t-test was conducted comparing AUROC scores with and without this label included. The scores for the model with 14 original labels (without “Uncertain”) and 15 labels (with “Uncertain”) showed minor variations, with mean AUROC values of 0.890 and 0.892 on all 14 pathology labels, respectively. The test suggests that the “Uncertain” label does not have a significant impact on the overall performance (five labels: $p = 0.0976$, 14 labels: $p = 0.2076$) of the model. However, the results indicate that adding this label leads to marginally better AUROCs, which may hold clinical importance when handling ambiguous cases.

\subsection{Penalty}
To assess the impact of hierarchical label grouping and penalty strategies, we conducted experiments with one primary hierarchy in Fig~\ref{fig:H4} and two types of penalties, fixed and data-driven, across a range of scale factors. The fixed penalty imposed a constant penalty for parent-child inconsistencies, while the data-driven penalty was based on the conditional probability of child labels given parent labels.
Table~\ref{table:high_level_comparison} presents the AUROC values across different hierarchical training strategies focused on high-level pathological categories, showcasing the impact of the two penalty types at various scale factors. The primary goal of this analysis is to examine how penalty types and scale factors influence classification performance across parent categories.
Data-driven paradigm showed significantly better results for Missing Lung Tissue and Opacity in high-level ($p = 0.0351$, $0.0003$, respectively), and Atelectasis, Edema, and Lung Opacity in low-level labels ($p = 0.0038$, $0.0225$, $0.0001$, respectively). However, no significant difference was found in other pathologies and mean AUROCs. Based on these results, we followed the study with the data-driven penalty approach at a scale factor of 0.5 that achieved the highest mean of AUROC (0.903), suggesting a favorable balance of penalties for boosting classifier performance on the high-level categories compared to other configurations. Additionally, it attains the highest AUROC values for Abnormal (0.942), Fluid Accumulation (0.922), and Other (0.903), indicating its strength in enhancing the model’s recognition of these specific classes.

\begin{table*}[ht]
    \centering
    \begin{tabular}{l c | c c c c c c c c }
        \hline
        \textbf{Penalty Type} & \textbf{Scale Factor} & \textbf{Abnormal} & \textbf{Cardiac} & \textbf{Fluid Accumulation} & \textbf{Missing Lung Tissue} & \textbf{Opacity} & \textbf{Other} & \textbf{Mean} \\
        \hline
        Data-Driven  & 0.3  & 0.922 & 0.849 & 0.913 & \textbf{[0.875]} & \textbf{[0.933]} & 0.847 & 0.890 \\
        Data-Driven  & 0.5 & \textbf{[0.942]} & 0.857 & \textbf{[0.922]} & 0.873 & 0.929 & \textbf{[0.903]} &\textbf{[0.904]} \\
        Data-Driven  & 0.7 & 0.932 & \textbf{0.858} & 0.912 & 0.872 & 0.930 & 0.875 & 0.897 \\
        Data-Driven  & 1.0 & 0.941 & 0.854 & 0.918 & 0.868 & 0.932 & 0.888 & 0.900 \\
        \hline
        Fixed  & 0.3 & \textbf{0.935} & \textbf{[0.862]} & \textbf{0.917} & \textbf{0.871} & 0.923 & \textbf{0.900} & \textbf{0.901} \\
        Fixed  & 0.5 & 0.928 & 0.846 & 0.910 & 0.863 & \textbf{0.924} & 0.861 & 0.889 \\
        Fixed  & 0.7 & 0.931 & 0.854 & 0.916 & 0.863 & 0.921 & 0.851 & 0.889 \\
        Fixed  & 1.0 & 0.917 & 0.849 & 0.916 & 0.865 & 0.923 & 0.843 & 0.886 \\
        \hline
    \end{tabular}
    \caption{AUROC Comparison for Hierarchical Training Strategy on High-Level Categories Evaluated on the Official CheXpert Test Set. The highest AUROC is highlighted in \textbf{BOLD} for each column and penalty type. The Overall Maximum for each column is in brackets (\textbf{[*]}). All scenarios include the ``Uncertain'' label. $\beta = 1$ for all fixed penalty strategies.}
    \label{table:high_level_comparison}
\end{table*}

\subsection{Overall Model Performance}
\begin{figure}[ht]
    \centering
    \includegraphics[width=0.9\columnwidth]{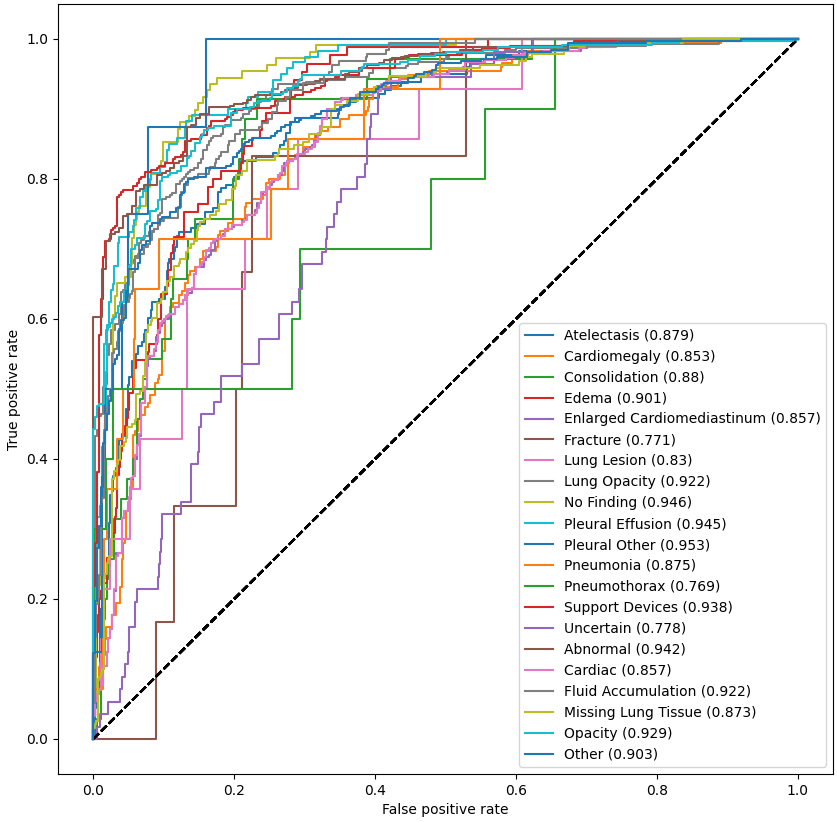}
    \caption{AUROC curves for the Hierarchical training strategy with a Data-Driven Penalty and a scale factor of 0.5 on the CheXpert dataset.}
    \label{fig:AUROC_H4}
\end{figure}

Overall, our model achieves a mean AUROC of 0.903, 0.904 (see table~\ref{table:high_level_comparison}), and 0.892 (see table~\ref{table:abalation}) on all the hierarchy, the high-level, and the five common CheXpert labels, respectively. As depicted in Fig~\ref{fig:AUROC_H4}, the AUROC curves show the model's overall efficacy in accurately distinguishing between pathologies and its ability to generalize well across diverse CXR features, supporting its applicability in clinical settings where accurate multi-label classification is essential.

High AUROC values for high-level classes suggest the model’s capacity to accurately identify broader pathologies, an important requirement in clinical applications. In contrast, it highlights the challenges in detecting specific pathologies that may have overlapping visual characteristics or subtle manifestations on CXRs.

\subsection{Visualization}
\begin{figure*}[ht]
    \centering
    \includegraphics[width=\textwidth]{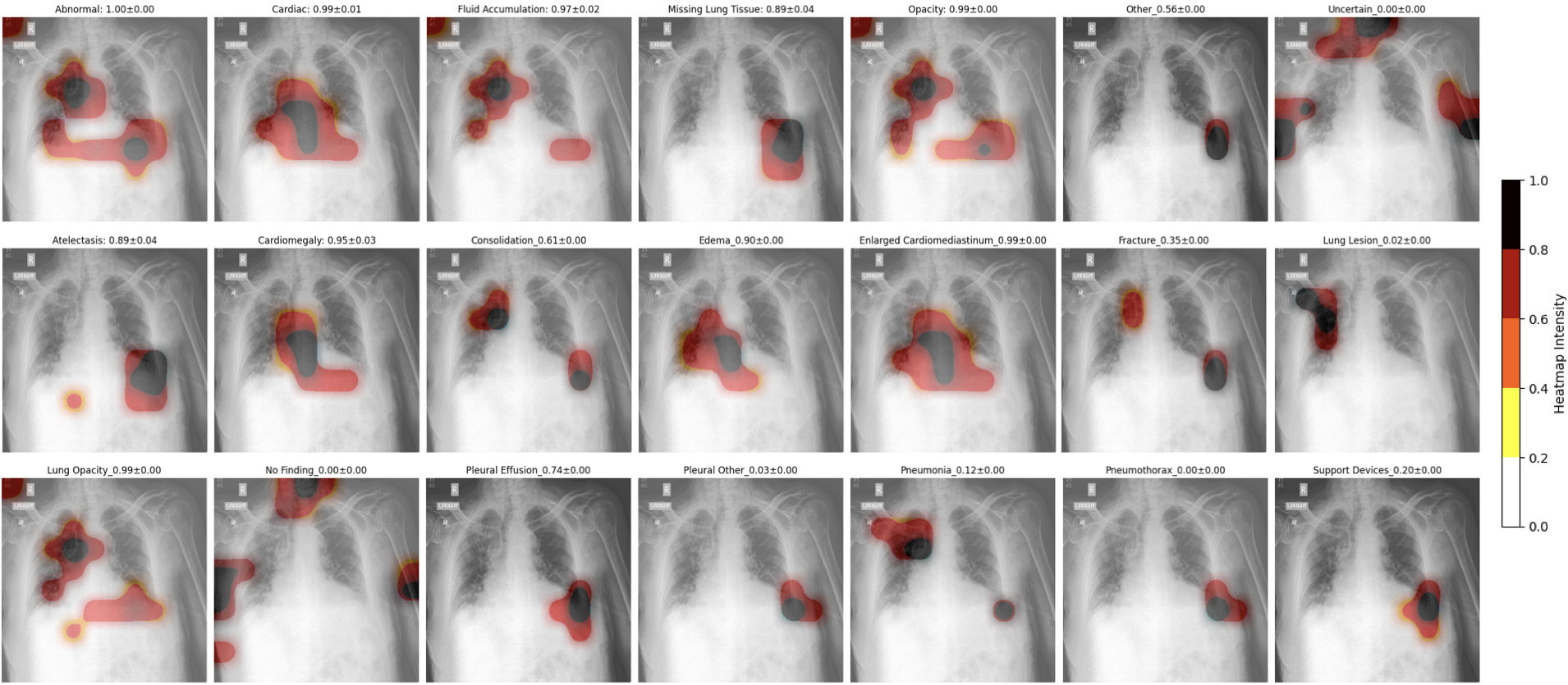}
    \caption{Activations on the sample test image. \newline Top row: defined high-level and ``Uncertain'' labels. Middle and bottom rows: original labels in the CheXpert dataset.\newline Activations are normalized and clipped with a 0.5 threshold, and the color map is segmented to enhance both localization and interpretation.\newline In this example, Fluid Accumulation shows high activation over potential fluid-affected areas, consistent with findings in Pleural Effusion and Edema. In the bottom row, child pathologies such as Cardiomegaly, Lung Opacity, and Enlarged Cardiomediastinum show targeted focus in the heart region and other expected locations, validating the model’s spatial attention. \newline Prediction confidence and uncertainty for ${N=10}$ Monte Carlo iterations can be found on top of each sub-image.\newline \textbf{Ground truth:} Atelectasis, Cardiomegaly, Edema, Enlarged Cardiomegaly, Lung Opacity, Pleural Effusion}
    \label{fig:sample_activation}
\end{figure*}

Fig~\ref{fig:sample_activation} depicts generated CAMs, where each sub-image corresponds to a specific label, showing the model’s activation regions overlaid on the original CXR. In these CAMs, the calculated activations from the gradients are clipped after normalization, with a 0.5 threshold to retain only the most prominent regions. This technique effectively filters out lower activation values, focusing the visualization on the areas of highest relevance to each pathology. Additionally, the discretized colormap segments the color spectrum to enhance the clarity of localized focus. The color bar in the Figure indicates the activation intensity, with values close to 1 (black) representing areas of high model attention and values near 0 (white) showing minimal focus. This segmentation provides a more distinct contrast between high and low activation areas, allowing for better interpretability and enabling clearer identification of regions the model attributes greater attention.

Together, clipping the activations and segmented colormap approach aims to highlight key areas on the CXR where the model “looks” to predict each pathology, making the activation maps more clinically meaningful by limiting visual noise and emphasizing the most diagnostically relevant areas.

\section{Discussion}
The goal of this work was to create a highly accurate model to classify CXR diagnoses for clinical decision support. The hierarchical structure of the model is a key strength that enhances interpretability by aligning label dependencies with medical expertise and categorization. This aims to improve prediction accuracy, especially in complex, multi-label settings like CXR analysis. Additionally, the custom hierarchical loss function, which applies penalties for inconsistent parent-child predictions, ensures logical consistency between predictions. 

A strength of the method is the single-run training pipeline, which simplifies deployment and reduces computational overhead, making the approach practical for real-world clinical applications. The integration of CAMs and Monte Carlo uncertainty estimates also enhances transparency, providing both interpretability and confidence measures for clinicians.

In comparison to previous works (see Table~\ref{table:abalation}), our proposed hierarchical model achieves competitive performance across all pathologies, recording the highest AUROC for Atelectasis (0.879) and Pleural Effusion (0.945), while maintaining a mean AUROC (0.892) similar to other state-of-the-art studies. Our method performed marginally better than Zhang \textit{et al.}~\cite{zhang2024discriminative} ($p = 0.0598$), however, no significant difference was found between all these methods. 

\begin{table}[!ht]
    \centering
    \begin{tabular}{l | c c c c c c c}
        \hline
        \textbf{Methods} &\textbf{Ate} & \textbf{Car} & \textbf{Con} & \textbf{Ede} & \textbf{Eff} & \textbf{Mean} \\
        \hline
        Wang \textit{et al.}~\cite{wang2024awareness} & 0.741 & 0.864 & 0.708 & 0.847 & 0.890 & 0.810  \\
        Zhang \textit{et al.}~\cite{zhang2024discriminative} & 0.667 & 0.850 & 0.713 & 0.829 & 0.862 & 0.784 \\
        Lu \textit{et al.}~\cite{lu2024CvTGNet} & 0.731 & \textbf{0.900} & 0.784 & 0.864 & 0.881 & 0.832 \\
        Liu \textit{et al.}~\cite{liu2023Global} & 0.811 & 0.853 & \textbf{0.943} & 0.912 & 0.928 & 0.889 \\
        Chen \textit{et al.}~\cite{chen2022Graph} & 0.747 & 0.882 & 0.785 & 0.888 & 0.916 & 0.844 \\
        Irvin \textit{et al.}~\cite{irvin2019chexpert} & 0.811 & 0.840 & 0.932 & 0.929 & 0.931 & 0.889 \\ 
        \textbf{*Pham \textit{et al.}}~\cite{Pham2021hierarchical} & 0.806 & 0.833 & 0.929 & \textbf{0.933} & 0.921 & 0.884  \\
        \hline
        \textbf{*Ours} & \textbf{0.879} & 0.853 & 0.880 & 0.901 & \textbf{0.945} & \textbf{0.892} \\
    \end{tabular}
    \caption{Comparison of AUROC scores between our proposed model and related studies for 5 common CheXpert pathologies. For a fair comparison, we put results for U-zeros for each study unless not stated. 
    The highest AUROC is highlighted in \textbf{BOLD} for each column. \textbf{Hierarchical} methods are in \textbf{bold} and marked with a \textbf{*}. Ate = Atelectasis, Car = Cardiomegaly, Con = Consolidation, Ede = Edema, Eff = Pleural Effusion.}
    \label{table:abalation}
\end{table}

In terms of limitations, the batch size affects the model performance by balancing generalization, convergence, and label representation. Smaller batches improve generalization but introduce noisy gradients, while larger batches stabilize training at the cost of higher computational demands and potential overfitting. Moreover, the model’s reliance on a specific label structure reduces its generalizability across datasets.

\section{Conclusion}
This study aimed to enhance the interpretability and clinical relevance of multi-label classification for CXR analysis by introducing a hierarchical label structure and a custom loss function. We organized the pathologies into clinically meaningful categories that reflect real-world clinical decision-making, enabling the model to capture the relationships and dependencies between various conditions more effectively. Additionally, by providing visual heatmaps of the activation areas driving our model's diagnostic decisions, we enable clinicians to correlate clinical and radiographic findings with the AI model for consistency.

We implemented the HBCE loss function that incorporates penalties for inconsistencies between parent and child predictions. This approach allowed for flexible experimentation with both fixed and data-driven penalty schemes, revealing their impact on model performance. The results demonstrated that using data-driven penalties potentially improves the model’s AUROC, indicating the benefit of penalizing clinically inconsistent predictions.

Our proposed hierarchical classification framework was evaluated across the hierarchy and multiple penalty configurations, showing that a single model with a single-run training pipeline could achieve a weighted AUROC of 0.9034 on the CheXpert dataset. The framework’s interpretability was further enhanced using CAM visualizations and Monte Carlo uncertainty estimation, which provided insights into the model’s decision-making process and its confidence in predictions.

Overall, the findings underline the value of integrating clinically inspired hierarchical structures and customized loss functions in medical image analysis. This work contributes to bridging the gap between automated classification and clinical application by promoting more interpretable and clinically aligned predictions.


Future work could explore extending this study to use adaptive batch size strategies to combine the benefits of small-batch generalization and large-batch stability, as well as incorporating a domain adaptation approach that dynamically aligns the hierarchy structure to the target dataset. The model's impact should also be evaluated in simulated clinical scenarios, such as Objective Structured Clinical Examinations (OSCEs), to validate its effectiveness as a clinical decision-support tool in real-world workflows.

\section*{Acknowledgment}
The following work was funded by an NSERC Discovery Grant (RGPIN-2017-06722).


\begin{thebibliography}{00}

\bibitem{Dattani2023Causes}
S. Dattani, F. Spooner, H. Ritchie, and M. Roser, 
``Causes of Death,'' 
\textit{Our World in Data}, 2023. 
Available: \href{https://ourworldindata.org/causes-of-death}{https://ourworldindata.org/causes-of-death}.

\bibitem{Kasalak2023Workload}
\"{O}. Kasalak, H. Alnahwi, R. Toxopeus, J. P. Pennings, D. Yakar, and T. C. Kwee, 
``Work overload and diagnostic errors in radiology,'' 
\textit{European Journal of Radiology}, vol. 167, p. 111032, 2023. 
doi: \href{https://doi.org/10.1016/j.ejrad.2023.111032}{10.1016/j.ejrad.2023.111032}.

\bibitem{Bruls2020Workload}
R. J. M. Bruls and R. M. Kwee, 
``Workload for radiologists during on-call hours: dramatic increase in the past 15 years,'' 
\textit{Insights into Imaging}, vol. 11, p. 121, 2020. 
doi: \href{https://doi.org/10.1186/s13244-020-00925-z}{10.1186/s13244-020-00925-z}.

\bibitem{irvin2019chexpert}
Irvin, Jeremy, Pranav Rajpurkar, Michael Ko, Yifan Yu, Silviana Ciurea-Ilcus, Chris Chute, Henrik Marklund, Behzad Haghgoo, Robyn Ball, Katie Shpanskaya, et al. \emph{CheXpert: A Large Chest Radiograph Dataset with Uncertainty Labels and Expert Comparison}.
In Proceedings of the Thirty-Third AAAI Conference on Artificial Intelligence, 2019.

\bibitem{jain2021visualchexbert}
Jain, Saahil, Akshay Smit, Steven QH Truong, Chanh DT Nguyen, Minh-Thanh Huynh, Mudit Jain, Victoria A Young, Andrew Y Ng, Matthew P Lungren, and Pranav Rajpurkar. \emph{VisualCheXbert: Addressing the Discrepancy Between Radiology Report Labels and Image Labels}.
arXiv preprint arXiv:2102.11467, 2021.

\bibitem{smit2020chexbert}
Smit, Akshay, Saahil Jain, Pranav Rajpurkar, Anuj Pareek, Andrew Y. Ng, and Matthew P. Lungren. \emph{CheXbert: Combining Automatic Labelers and Expert Annotations for Accurate Radiology Report Labeling Using BERT}.
arXiv preprint arXiv:2004.09167, 2020.

\bibitem{Huang2017DenseNet}
Huang, Gao, Zhuang Liu, Laurens van der Maaten, and Kilian Q. Weinberger. \emph{Densely connected convolutional networks.}
In *Proceedings of the IEEE Conference on Computer Vision and Pattern Recognition*, 2017.

\bibitem{selvaraju2017gradcam}
R.~R.~Selvaraju, M.~Cogswell, A.~Das, R.~Vedantam, D.~Parikh, and D.~Batra, ``Grad-CAM: Visual explanations from deep networks via gradient-based localization,'' in \emph{Proceedings of the IEEE International Conference on Computer Vision (ICCV)}, 2017, pp.~618--626. doi{10.1109/ICCV.2017.74}

\bibitem{wang2024awareness}
G. Wang, P. Wang, and B. Wei, "Multi-label local awareness and global co-occurrence priori learning improve chest X-ray classification," \textit{Multimedia Systems}, vol. 30, no. 132, pp. 1-12, 2024. doi: 10.1007/s00530-024-01321-z.

\bibitem{zhang2024discriminative}
K. Zhang, W. Liang, P. Cao, X. Liu, J. Yang, and O. Zaiane, "Label correlation guided discriminative label feature learning for multi-label chest image classification," \textit{Computer Methods and Programs in Biomedicine}, vol. 245, Article 108032, 2024. doi: 10.1016/j.cmpb.2024.108032.

\bibitem{lu2024CvTGNet}
Y. Lu, Y. Hu, L. Li, Z. Xu, H. Liu, H. Liang, and X. Fu, "CvTGNet: A Novel Framework for Chest X-Ray Multi-label Classification," \textit{Proceedings of the 21st ACM International Conference on Computing Frontiers (CF '24)}, May 2024, pp. 1-9. doi: 10.1145/3649153.3649216.

\bibitem{liu2023Global}
Z. Liu, Y. Cheng, and S. Tamura, "Multi-Label Local to Global Learning: A Novel Learning Paradigm for Chest X-Ray Abnormality Classification," \textit{IEEE Journal of Biomedical and Health Informatics}, vol. 27, no. 9, pp. 4409-4420, Sep. 2023. doi: 10.1109/JBHI.2023.3281466.

\bibitem{chen2022Graph}
B. Chen, Z. Zhang, Y. Li, G. Lu, and D. Zhang, "Multi-Label Chest X-Ray Image Classification via Semantic Similarity Graph Embedding," \textit{IEEE Transactions on Circuits and Systems for Video Technology}, vol. 32, no. 4, pp. 2455-2467, Apr. 2022. doi: 10.1109/TCSVT.2021.3079900.

\bibitem{ke2021CheXtransfer}
A. Ke, W. Ellsworth, O. Banerjee, A. Y. Ng, and P. Rajpurkar, "CheXtransfer: Performance and Parameter Efficiency of ImageNet Models for Chest X-Ray Interpretation," in \textit{Proceedings of the ACM Conference on Health, Inference, and Learning (CHIL)}, Virtual Event, 2021, pp. 2097-2106. doi: 10.1145/3450439.3451867.

\bibitem{huang2022Transfer}
G.-H. Huang, Q.-J. Fu, M.-Z. Gu, N.-H. Lu, K.-Y. Liu, and T.-B. Chen, "Deep Transfer Learning for the Multilabel Classification of Chest X-ray Images," \textit{Diagnostics}, vol. 12, no. 6, p. 1457, 2022. doi: 10.3390/diagnostics12061457.

\bibitem{wang2017ChestX}
X. Wang, Y. Peng, L. Lu, Z. Lu, M. Bagheri, and R. M. Summers, "ChestX-ray8: Hospital-scale chest X-ray database and benchmarks on weakly-supervised classification and localization of common thorax diseases," \textit{Proceedings of the IEEE Conference on Computer Vision and Pattern Recognition (CVPR)}, 2017, pp. 2097-2106.

\bibitem{Wehrmann2018Hierarchical} J. Wehrmann, R. Cerri, and R. C. Barros, ``Hierarchical Multi-Label Classification Networks,'' in \emph{Proceedings of the 35th International Conference on Machine Learning}, Stockholm, Sweden: PMLR, vol. 80, 2018, pp. 5075–5084.

\bibitem{Chen2020hierarchical} H. Chen, S. Miao, D. Xu, G. D. Hager, and A. P. Harrison, "Deep hierarchical multi-label classification applied to chest X-ray abnormality taxonomies," \emph{Medical Image Analysis}, vol. 66, 2020, Art. no. 101811.

\bibitem{Pham2021hierarchical} H. H. Pham, T. T. Le, D. Q. Tran, D. T. Ngo, and H. Q. Nguyen, "Interpreting chest X-rays via CNNs that exploit hierarchical disease dependencies and uncertainty labels," \emph{Neurocomputing}, vol. 437, pp. 186–194, 2021.

\bibitem{johnson2019MIMIC}
A. E. W. Johnson, T. J. Pollard, S. J. Berkowitz, et al., "MIMIC-CXR, a de-identified publicly available database of chest radiographs with free-text reports," \textit{Scientific Data}, vol. 6, Article 317, 2019. [Online]. Available: https://doi.org/10.1038/s41597-019-0322-0.

\bibitem{Aberle2000plco}
A. P. Aberle, M. D. Adams, C. J. Berg, et al., "The Prostate, Lung, Colorectal and Ovarian (PLCO) Cancer Screening Randomized Controlled Trial," \textit{Journal of the National Cancer Institute}, vol. 92, no. 18, pp. 1536-1544, 2000. [Online]. Available: https://cdas.cancer.gov/plco/

\bibitem{padchest2020}
A. Bustos, A. Pertusa, J. Salinas, and M. de la Iglesia-Vayá, "PadChest: A large chest x-ray image dataset with multi-label annotated reports," \textit{Medical Image Analysis}, vol. 66, 101797, 2020. [Online]. Available: https://doi.org/10.1016/j.media.2020.101797.

\bibitem{raghu2019transfusion}
M.~Raghu, C.~Zhang, J.~Kleinberg, and S.~Bengio, ``Transfusion: Understanding transfer learning for medical imaging,'' in \emph{Proceedings of the 33rd International Conference on Neural Information Processing Systems (NeurIPS)}, 2019, pp. 3342--3352.

\bibitem{smith2018disciplined}
Smith, L. N., "A disciplined approach to neural network hyper-parameters: Part 1--learning rate, batch size, momentum, and weight decay," *arXiv preprint arXiv:1803.09820*, 2018.

\bibitem{morton2024ai}
W.~Morton, ``{AI lifts nonradiologists' x-ray interpretation skill levels},'' \emph{AuntMinnie.com}, Oct. 25, 2024. [Online]. Available: \url{https://www.auntminnie.com/clinical-news/digital-x-ray/article/15706835/ai-lifts-nonradiologists-xray-interpretation-skill-levels}.

\bibitem{sen2021hierarchical}
C.~Sen, B.~Ye, J.~Aslam, and A.~Tahmasebi, ``From extreme multi-label to multi-class: A hierarchical approach for automated ICD-10 coding using phrase-level attention,'' \emph{arXiv preprint arXiv:2102.09136}, 2021. Available: \url{https://arxiv.org/abs/2102.09136}

\bibitem{webmd_pleural_effusion}
WebMD, ``Pleural Effusion: Symptoms, Causes, Treatments,'' [Online]. Available: \url{https://www.webmd.com/lung/pleural-effusion-symptoms-causes-treatments}.

\bibitem{mayo_clinic_pneumonia}
Mayo Clinic, ``Pneumonia,'' [Online]. Available: \url{https://www.mayoclinic.org/diseases-conditions/pneumonia/symptoms-causes/syc-20354204}. 

\bibitem{cleveland_clinic_pulmonary_edema}
Cleveland Clinic, ``Pulmonary Edema,'' [Online]. Available: \url{https://my.clevelandclinic.org/health/diseases/17841-pulmonary-edema}.

\bibitem{radiopaedia_}
Radiopaedia.org, ``Consolidation,'' [Online]. Available: \url{https://radiopaedia.org/articles/consolidation}.

\bibitem{smithuis2014lungdisease} R. Smithuis, “Chest X-Ray - Lung Disease,” The Radiology Assistant, 2014. [Online]. Available: \url{https://radiologyassistant.nl/chest/chest-x-ray/lung-disease}.

\end{thebibliography}
\end{document}